\pdfoutput=1
\documentclass[11pt]{article}

\usepackage[preprint]{acl}

\usepackage{times}
\usepackage{latexsym}
\usepackage{amsmath}
\usepackage[T1]{fontenc}
\usepackage[utf8]{inputenc}
\usepackage{algorithmic}  % 或 algorithmicx
\usepackage{xcolor}
\usepackage{xspace}

\usepackage{microtype}
\usepackage{amssymb}
\usepackage{inconsolata}

\usepackage{graphicx}

\usepackage{booktabs, multirow} % for borders and merged ranges
\usepackage{soul}% for underlines
\usepackage{changepage,threeparttable} % for wide tables

\newcommand{\methodname}{\textsc{DIVE}\xspace}

\title{\methodname: Diversified Iterative Self-Improvement}

\author{
Yiwei Qin$^{2}$ \and Yixiu Liu$^{1,2}$ \and Pengfei Liu$^{1,2}$\thanks{Corresponding author}\\[2mm]
$^1$Shanghai Jiao Tong University\\
$^2$Generative AI Research Lab (GAIR)\\[2mm]
\texttt{\{qinyiwei07@outlook.com, pengfei@sjtu.edu.cn\}}\\[1mm]
}

\begin{document}
\maketitle
\begin{abstract}
Recent advances in large language models (LLMs) have demonstrated the effectiveness of Iterative Self-Improvement (ISI) techniques. However, continuous training on self-generated data leads to reduced output diversity, a limitation particularly critical in reasoning tasks where diverse solution paths are essential. We present DIVE (Diversified Iterative Self-Improvement), a novel framework that addresses this challenge through two key components: Sample Pool Expansion for broader solution exploration, and Data Selection for balancing diversity and quality in preference pairs. Experiments on MATH and GSM8k datasets show that DIVE achieves a 10\% to 45\% relative increase in output diversity metrics while maintaining performance quality compared to vanilla ISI. Our ablation studies confirm both components' significance in achieving these improvements. Code is available at \url{https://github.com/qinyiwei/DIVE}.

%Recent advances in large language models have demonstrated the effectiveness of Iterative Self-Improvement (ISI) techniques. However, these approaches suffer from model collapse, manifesting as reduced output diversity. This limitation is particularly critical in reasoning tasks, where diverse solution paths are essential. We present DIVE (Diversified Iterative Self-Improvement), a novel framework that addresses this challenge through two key components: Sample Pool Expansion for broader solution exploration, and Data Selection with outlier detection and greedy selection algorithms for curating diverse preference pairs. Experiments on MATH and GSM8k datasets show that DIVE achieves a 10\% to 45\% relative increase in output diversity metrics while maintaining performance quality compared to vanilla ISI. Our ablation studies confirm both components' significance in achieving these improvements. Code is available at \url{https://github.com/qinyiwei/DIVE}.

%Preference learning with an iterative approach has become a crucial method for aligning Large Language Models (LLMs) with human preferences. However, current mainstream methods overlook XX, leading to a decline in model performance in the YY aspect. To address this issue, we propose a simple method ZZ ...

\end{abstract}

\section{Introduction}
\begin{figure*}[t]
\centering
\includegraphics[width=\linewidth]{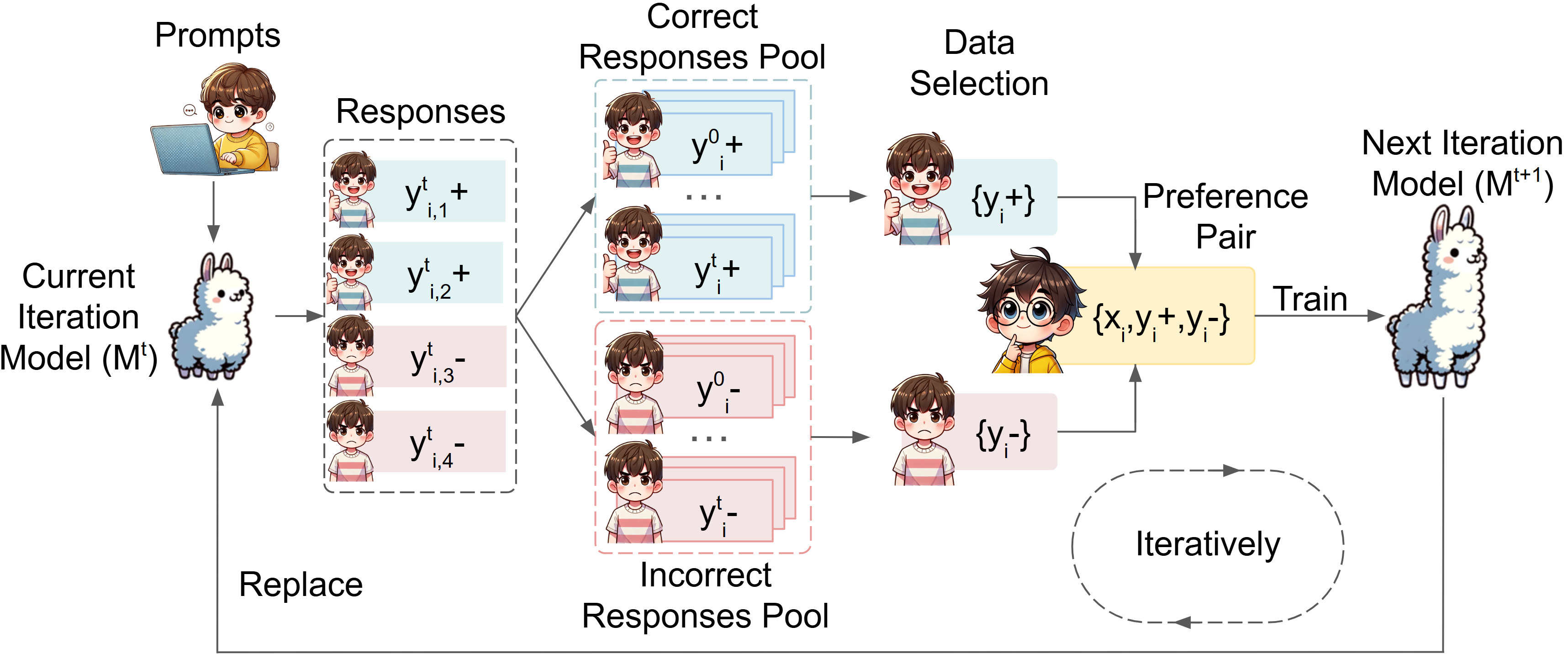}
\caption{Overview of the Diversified Iterative Self-Improvement (\methodname) framework. At each iteration $t$, the process includes response generation, pool expansion through correct and incorrect response collection, data selection for balancing quality and diversity, and model refinement through preference learning, producing an improved model $M^{t+1}$ for the next iteration.} 
\label{fig: model}
\end{figure*}

Recent advancements in large language models (LLMs) have driven significant improvements through self-improvement techniques~\citep{zelikman2022star,madaan2024self,wang2022self}, where models enhance their capabilities by refining their performance based on feedback, often using their own outputs for further enhancement. Two prominent approaches in this area are Reinforcement Learning (RL)~\citep{christiano2017deep, ziegler2019fine, ouyang2022traininglanguagemodelsfollow} and Preference Learning~\citep{rafailov2024direct, zhao2023slic, ethayarajh2024kto, azar2024general}, both of which enable models to refine their behavior by optimizing for feedback signals, such as rewards or preferences. Iterative Self-Improvement (ISI) extends these methods by using an iterative process, where models continuously leverage previous outputs to generate more refined responses, proving highly effective in various domains from general instruction-following~\citep{xu2023some,DBLP:conf/icml/YuanPCLSXW24} to specialized areas like mathematical reasoning~\citep{DBLP:journals/corr/abs-2404-19733, mitra2024orca}.

%Despite the positive outcomes of ISI in enhancing model performance, recent research has highlighted a significant challenge: model collapse~\citep{shumailov2024curserecursiontraininggenerated,dohmatob2024taletailsmodelcollapse,gerstgrasser2024modelcollapseinevitablebreaking}. This phenomenon occurs when models over-optimize for high-reward outputs, leading to reduced diversity in generated responses. This issue becomes especially pronounced in self-improvement settings, where models continuously refine their behavior by learning from their own outputs. Techniques like RL and Preference Learning, which aim to optimize for high-reward or preferred responses, exacerbate this problem by encouraging the model to focus on a limited set of solutions, ultimately reducing the variability of the model's outputs over time~\citep{DBLP:journals/corr/abs-2407-05013,kirk2023understanding}.

Despite the positive outcomes of ISI in enhancing model performance, recent research has identified model collapse as a critical challenge when training models on self-generated data~\citep{shumailov2024curserecursiontraininggenerated,dohmatob2024taletailsmodelcollapse,gerstgrasser2024modelcollapseinevitablebreaking}. This phenomenon, where models progressively lose information about the underlying distribution, is particularly relevant to ISI processes as models continuously learn from their own outputs. In RL and preference learning settings, this issue manifests as reduced diversity in generated responses, as the model increasingly focuses on a narrow set of high-reward patterns~\citep{DBLP:journals/corr/abs-2407-05013,kirk2023understanding}.

While recent advancements in reasoning with LLMs have focused on improving accuracy through top-ranking solutions, they often overlook the importance of diverse reasoning paths. Methods like Self Consistency~\citep{wang2022self}, ToT~\citep{yao2024tree} and RAP~\citep{hao2023reasoning} rely on the LLM's capacity to explore diverse reasoning solutions, leveraging the intuition that complex reasoning tasks typically admit multiple valid paths to the correct answer~\citep{evans2010intuition,stanovich2012distinction}. Although some studies have investigated techniques to enhance reasoning diversity \citep{wang2022self,xie2024self,li2022making,naik2023diversity,yu2024flow}, the challenge of diversity loss in ISI remains underexplored.

To address this challenge, we present \textbf{D}iversified \textbf{I}terative Self-Impro\textbf{VE}ment (\methodname), shown in Fig.\ref{fig: model}, the first study focused on this problem. \methodname operates through two complementary strategies in the preference learning stage: (1) Sample Pool Expansion and (2) Data Selection. Sample Pool Expansion encourages the model to explore a broader set of potential solutions at each iteration by sampling more responses per question and incorporating data from all previous iterations. Data Selection then applies outlier detection techniques to filter responses for quality while using greedy selection algorithms to maximize diversity in the preference pairs. By curating diverse yet high-quality preference pairs, \methodname guides the model to generate varied outputs while maintaining performance.

Our experimental results demonstrate that \methodname significantly enhances the diversity of model outputs on the MATH and GSM8k datasets compared to vanilla ISI, achieving a 10\% to 45\% relative increase across various diversity metrics for both positive and negative examples, without compromising output quality. Ablation studies further highlight the critical roles of Sample Pool Expansion and Data Selection in driving these results.

%The proposed \methodname approach represents a significant step towards developing more robust and versatile reasoning models by balancing diversity and performance in ISI. The techniques introduced in this study have the potential to extend to a wide range of tasks beyond reasoning, opening up new avenues for future research in self-improving AI systems. By addressing the challenge of diversity loss in ISI, \methodname contributes to the advancement of LLMs capable of generating diverse and high-quality outputs, ultimately enhancing their adaptability and generalizability in real-world applications.

\section{Methodology}

% formulate preference learning
% formulate iterative preference learning
% mention the tasks we aim to evaluate in this work

%Let $D = {(x_i, y_i)}_{i=1}^N$ represent a training set containing questions $x_i$ and their corresponding ground truth response $y_i$. We begin with a foundation model, typically a pre-trained model denoted as $M_{\text{PT}}$. While post-training generally involves both supervised fine-tuning (SFT) and preference learning, we focus on the preference learning stage where diversity reduction is most prominent, particularly in mathematical reasoning tasks. This focus allows us to directly address the core challenge of maintaining solution diversity while improving model performance through self-improvement.

Let $D = {(x_i, y_i)}_{i=1}^N$ represent a training set containing questions $x_i$ and their corresponding ground truth response $y_i$. We begin with a foundation model, typically a pre-trained model denoted as $M_{\text{PT}}$. 
%The post-training process generally involves two stages: supervised fine-tuning (SFT) and preference learning. While the concept of self-improvement could encompass the entire post-training workflow ~\cite{DBLP:journals/corr/abs-2407-05013}, in this paper, we focus specifically on the preference learning stage.
%The objective of self-improvement is to enhance the model's performance by refining its capabilities using its own outputs, without relying on external signals. When this process is repeated over multiple training rounds, it becomes ISI. In iterative self-improvement, the model incrementally improves by applying preference learning to its own generated responses at each iteration.
The objective of self-improvement is to enhance the $M_{\text{PT}}$'s performance by refining its capabilities using its own outputs, without relying on external signals. When this process is repeated over multiple training rounds, it becomes ISI, where the model incrementally improves by applying preference learning to its own generated responses at each iteration.

\subsection{Iterative Self Improvement}
\paragraph{Direct Preference Optimization (DPO) \citep{rafailov2024direct}}
DPO is a widely-used method for offline preference learning that enables direct optimization of model preferences without requiring an explicit reward model. The key insight of DPO is to express the probability of preference data using the ratio between the policy model and a reference model. The DPO objective is defined as:

% \begin{equation}
% L_{\text{DPO}}(\pi_\theta; \pi_{\text{ref}}) = - \mathbb{E}_{(x, y^+, y^-) \sim  D_{\text{pref}}}  \left[ \log \sigma \left( \beta \log \frac{\pi_\theta(y^+ | x)}{\pi_{\text{ref}}(y^+ | x)} - \beta \log \frac{\pi_\theta(y^- | x)}{\pi_{\text{ref}}(y^- | x)} \right) \right]

% \end{equation}

% \begin{equation}
% \begin{aligned}
% L_{\text{DPO}}(\pi_\theta; \pi_{\text{ref}}) = - \mathbb{E}_{(x, y^+, y^-) \sim  D_{\text{pref}}} & \left[ \log \sigma \left( \beta \log \frac{\pi_\theta(y^+ | x)}{\pi_{\text{ref}}(y^+ | x)} \right. \right. \\
% & \left. \left. - \beta \log \frac{\pi_\theta(y^- | x)}{\pi_{\text{ref}}(y^- | x)} \right) \right]
% \end{aligned}
% \end{equation}

% \begin{equation}
% \begin{split}
% & \quad L_{\text{DPO}}(\pi_\theta; \pi_{\text{ref}}) \\
% & =- \mathbb{E}_{(x, y^+, y^-) \sim  D_{\text{pref}}} \Big[ \log \sigma \big( \beta \log \frac{\pi_\theta(y^+ | x)}{\pi_{\text{ref}}(y^+ | x)} \\
% & \qquad\qquad - \beta \log \frac{\pi_\theta(y^- | x)}{\pi_{\text{ref}}(y^- | x)} \big) \Big]
% \end{split}
% \end{equation}

\begin{multline}
L_{\text{DPO}}(\pi_\theta; \pi_{\text{ref}}) = -\mathbb{E}_{(x, y^+, y^-) \sim D_{\text{pref}}} \left[ \log \sigma(r) \right], \\
r = \beta \log \frac{\pi_\theta(y^+ | x)}{\pi_{\text{ref}}(y^+ | x)} - \beta \log \frac{\pi_\theta(y^- | x)}{\pi_{\text{ref}}(y^- | x)}
\end{multline}

where $ (x, y^+, y^-) $ represents preference pairs from the preference dataset $ D_{\text{pref}}$, with $x $ being the input question, $y^+ $ the preferred (correct) response, and $y^- $ the non-preferred (incorrect) response. The policy model $\pi_\theta $ learns to assign higher probability to preferred responses compared to non-preferred ones.

To stabilize the DPO training and prevent the model from deviating too far from its initial behavior, we incorporate an additional negative log-likelihood (NLL) loss on the chosen sequences~\citep{DBLP:journals/corr/abs-2404-19733, dubey2024llama, xu2024chatglm}. This helps maintain response consistency while allowing for targeted improvements through preference learning. The NLL loss term is defined as:

 \begin{equation}
L_{\text{NLL}} = - \mathbb{E}_{(x, y^+) \sim  D_{\text{pref}}} \frac{\log \pi_\theta(y^+ | x)}{|y^+|}
 \end{equation}

The final loss function combines the DPO and NLL losses as follows:
 % \begin{equation}
 % L_{\text{pref} = \alpha \cdot L_{\text{DPO}} + (1 - \alpha) \cdot L_{\text{NLL}}
 % \end{equation}
 \begin{equation}
L_{\text{pref}} = \alpha \cdot L_{\text{DPO}} + (1 - \alpha) \cdot L_{\text{NLL}}
\end{equation}

where $\alpha $ is a hyperparameter that balances the contributions of DPO and NLL losses.

\paragraph{Iterative Training}
We start by performing supervised fine-tuning (SFT) on the pre-trained model $ M_{\text{PT}} $ using dataset $ D $, producing a fine-tuned model $ M_0 $. In ISI, a series of models $ M_1, \dots, M_T $ are trained, where each model $ M_t $ builds upon the outputs of the previous model $ M_{t-1} $. During each iteration, preference data for training $ M_t $ is sampled from $ M_{t-1} $, and $ M_{t-1} $ is used as the reference model in the DPO loss. The steps for each iteration are as follows:
\begin{enumerate}
\item \textbf{Data Sampling}: In the $ t $-th iteration, for each question $ x \in D $, we sample $ K $ responses from the model $ M_{t-1} $ to form the candidate pool: $ D_{\text{pool}}^t = \{(x_i, y_i^j) | x_i \in D, j \in [1, K]\} $.

\item  \textbf{Preference Pair Construction}: The candidate pool $ D_{\text{pool}}^t $ is divided into a correct pool $ D_{\text{pool}}^{t+} $ and an incorrect pool $ D_{\text{pool}}^{t-} $ by comparing the generated response with the gold-standard answer. If the final answer of a generated response matches the gold standard, the response goes to $ D_{\text{pool}}^{t+} $; otherwise, it goes to $ D_{\text{pool}}^{t-} $. From these pools, we select $ P $ responses to construct the preference dataset:

   $
   D_{\text{pref}}^t = \{(x_i, y_i^+, y_i^-) | x_i \in D, y_i^+ \in D_{\text{pool}}^{t+}, y_i^- \in D_{\text{pool}}^{t-}\}.
   $

\item  \textbf{Preference Training}: Using the preference dataset $ D_{\text{pref}}^t $, the model $ M_{t-1} $ is refined into $ M_t $ by optimizing the preference loss $ L_{\text{pref}} $.
\end{enumerate}

\subsection{Diversified Iterative Self-Improvement}
As highlighted in~\citet{DBLP:journals/corr/abs-2407-05013,kirk2023understanding}, preference learning often leads to a reduction in diversity, a problem that is exacerbated in iterative settings due to the accumulation of this effect over time. 
We propose two complementary strategies to address this challenge: Sample Pool Expansion, which enlarges the candidate pool for response selection, and Data Selection, which ensures diverse yet high-quality examples are chosen for training. These strategies work within the existing ISI framework while effectively maintaining output diversity.

\subsubsection{Sample Pool Expansion}
To provide more candidates for constructing diverse preference pairs, we expand the candidate sample pool $D_{\text{pool}}$ through two complementary strategies. A larger sample pool offers more options for the subsequent data selection process, which is crucial for selecting diverse examples for preference learning.

\paragraph{Increased Sampling per Question} At each iteration, we increase the number of responses K sampled per question, providing a broader set of candidates for preference learning.

\paragraph{Global Data Usage}
\label{sec:global-data}
Instead of relying solely on the responses generated by model $ M_{t-1} $ for training $ M_t $, we incorporate global data from all previous iterations. This expanded pool is defined as $ D_{\text{pool}}^t  = \bigcup_{i=1}^t  D_{\text{pool}}^i $ ensuring that no information from previous iterations is lost and avoiding extra sampling computation.

\subsubsection{Data Selection}
\label{sec:data-selection}
Our preliminary experiments show that the diversity of the examples selected for preference learning, rather than the overall diversity of the response pool, significantly impacts the model's ability to generate diverse outputs after training. Thus, it is crucial to carefully select diverse examples from the response pool for preference learning.

\paragraph{Greedy Selection Method}
We use a greedy algorithm to maximize the diversity of the selected responses, following these steps:

\begin{enumerate}
\item Randomly select one response from $ D_{\text{pool}}$ and add it to the selected response list. Remove this response from $ D_{\text{pool}}$.
\item For each remaining response in $ D_{\text{pool}}$, calculate the diversity of the selected response list as if the current example were added.
\item Select the response that maximizes the diversity of the selected list, add it to the list, and remove it from $ D_{\text{pool}}$.
\item Repeat Steps 2 and 3 until either $ D_{\text{pool}}=\varnothing$ or the desired number of responses P is reached.
\end{enumerate}
While this method increases diversity effectively, we observed that focusing solely on diversity can negatively impact model accuracy. We hypothesize that maximizing diversity may lead to selecting low-quality, outlier responses that harm the model's performance.

\paragraph{Balancing Quality and Diversity}
To mitigate this issue, we first filter the response pool using the Isolation Forest method \citep{4781136}, with features derived from Sentence-BERT embeddings \citep{reimers2019sentence} that capture the semantic aspects of the responses. Using distances in the embedding space, we identify and exclude extreme outliers (responses that deviate significantly from the general distribution of valid solutions) to maintain response quality. 

Once the response pool is filtered, we apply the greedy selection method to maximize diversity among the remaining high-quality responses. This ensures a balanced selection process that maintains both diversity and quality in the final model.

\section{Experiment}

\begin{table*}[!htp]\centering

\small
\begin{tabular}{lrrrrrrrr}\toprule
&Method &Dis-N Pos &Dis-N Neg &SentBERT Pos &SentBERT Neg &@1 &@50 \\\midrule
\multirow{4}{*}{Sample 10} &Vanilla &0.345 &0.454 &0.111 &0.168 &0.704 &0.976 \\
&Global &0.350 &0.444 &0.119 &0.182 &\underline{0.707}&\underline{\textbf{0.980}} \\
&Selection &0.388 &0.462 &0.125 &0.184 &0.703 &0.975 \\
&Global+Selection &\underline{0.397} &\underline{0.507} &\underline{0.132} &\underline{0.196} &\underline{0.707} &0.975 \\\midrule
\multirow{4}{*}{Sample 50} &Vanilla &0.309 &0.380 &0.106 &0.168 &0.718 &0.975 \\
&Global &0.348 &0.462 &0.118 &0.184 &0.716 &0.974 \\
&Selection &0.440 &\underline{\textbf{0.538}} &0.145 &0.214 &0.716 &\underline{0.976} \\
&Global+Selection &\underline{\textbf{0.448}} &0.502 &\underline{\textbf{0.152}} &\underline{\textbf{0.224}} &\underline{\textbf{0.722}} &0.972 \\
\bottomrule

\end{tabular}
\caption{Comparison of different diversity enhancement methods on GSM8k dataset using Mistral-7B as the base model. Results show diversity metrics (Dis-N and SentBERT) for both positive and negative examples, along with accuracy metrics. All metrics have been normalized so that higher values consistently indicate better performance. \textbf{Bold} indicates the best overall performance across all settings, while \underline{underline} represents the best performance within their respective sampling group (Sample 10 or Sample 50).}
\label{tab: exp_result_gsm8k}
\end{table*}

\begin{table*}[!htp]\centering

\small
\begin{tabular}{lrrrrrrrr}\toprule
&Method &Dis-N Pos &Dis-N Neg &SentBERT Pos &SentBERT Neg &@1 &@50 \\\midrule
\multirow{4}{*}{Sample 10} &Vanilla &0.647 &0.557 &0.247 &0.304 &0.176 &0.580 \\
&Global &0.636 &0.550 &0.242 &0.300 &\underline{\textbf{0.194}} &\underline{\textbf{0.610}} \\
&Selection &0.662 &0.565 &0.245 &\underline{0.311} &0.178 &0.600 \\
&Global+Selection &\underline{0.665} &\underline{0.573} &\underline{0.254} &0.310 &0.188 &\underline{\textbf{0.610}} \\\midrule
\multirow{4}{*}{Sample 50} &Vanilla &0.612 &0.540 &0.228 &0.283 &0.186 &\underline{0.606} \\
&Global &0.635 &0.542 &0.247 &0.299 &0.190 &\underline{0.606} \\
&Selection &\underline{\textbf{0.694}} &\underline{\textbf{0.612}} &0.264 &0.313 &0.188 &0.594 \\
&Global+Selection &0.692 &0.599 &\underline{\textbf{0.273}} &\underline{\textbf{0.326}} &\underline{\textbf{0.194}} &0.586 \\
\bottomrule
\end{tabular}
\caption{Results on MATH dataset with identical experimental settings as Table~\ref{tab: exp_result_gsm8k}.}\label{tab: exp_result_math}
\end{table*}

\subsection{Experimental Settings}
\subsubsection{datasets}
We conducted experiments on two math reasoning datasets: 

\paragraph{GSM8K} \citep{cobbe2021training}: This dataset contains grade-school-level math word problems. Each problem consists of a question $x_i$ and a solution  $y_i$, which includes a gold chain-of-thought (COT) explanation~\citep{wei2022chain} and a final numerical answer. The training set consists of 7,473 examples, and the test set contains 1,319 examples.

\paragraph{MATH} \citep{hendrycks2021measuring}: This dataset contains more advanced math problems. Similar to GSM8K, each problem provides a gold CoT solution along with a final answer. The training set includes 7,500 problems, while the test set contains 5,000 examples.

In the self-improvement paradigm, for both datasets, we utilize only the questions from the training set for preference learning, without introducing any additional questions. The correctness of the model-generated solutions is judged based on the final answers provided in the gold solutions.

\subsubsection{Evaluation Metrics}\label{sec: evaluation metric}
To assess how well the model balances quality and diversity, we adopt two types of evaluation metrics that measure performance from both aspects:

\paragraph{Quality} 

For quality evaluation, we use the following metrics: \textbf{@1 Accuracy} which measures the model's accuracy when sampling a single response. It tests how well the model ranks the sample space, with a focus on whether the correct response is placed at the top-1 position. \textbf{@50 Accuracy} which evaluates the model's accuracy when sampling 50 responses. The model is considered correct if any of the 50 responses is correct. This metric tests the model's potential to solve a question when sampling more responses.

\paragraph{Diversity}

To evaluate the diversity of the generated responses, we use the following metrics, in line with \citet{kirk2023understanding}: \textbf{Distinct N-grams}~\citep{tevet2020evaluating} which counts the number of distinct N-grams (averaged over $ n = 1, \dots, 5 $) in the set of outputs, which provides a measure of lexical diversity. \textbf{Sentence-BERT Embedding Cosine Similarity}~\citep{li2015diversity} which embeds each response using a Sentence-BERT model and calculates the average cosine similarity between the embeddings. The diversity score is then calculated as $ 1 - \text{average similarity} $, where lower similarity indicates higher diversity. Both of these methods have been shown to align well with human evaluations of diversity \citep{tevet2020evaluating}, enabling us to quantify the diversity of the model's outputs effectively.

\subsubsection{Training Details}
\label{sec:training-details}
Our experiments are based on the pre-trained language model Mistral-7B \citep{jiang2023mistral}. For SFT, we fine-tune Mistral-7B on the GSM8K/MATH Train subset to produce the initial model, $M_0$. The fine-tuning is done using full-model fine-tuning with a learning rate of $1 \times 10^{-6}$, a cosine learning rate schedule, 3 epochs.

For the ISI phase, at each iteration $t$, we generate $K = 10$ or $50$ solutions per question from the GSM8K/MATH Train subset to form the response pool $D_{\text{pool}}^t$, using nucleus sampling with $ \text{top}\_p = 0.95 $ and temperature $ T = 0.7 $, based on the model $M_{t-1}$. For experiments without global data usage, $P = 5$ preference pairs are constructed from $D_{\text{pool}}^t$. For experiments with global data usage, the pool is expanded to $D_{\text{pool}}^t = \cup_{i=1}^{t} D_{\text{pool}}^i$. \footnote{Since some questions may have fewer than $P = 5$ correct or incorrect responses, we construct at most P preference pairs per question. Questions with no correct or no incorrect responses in the pool are skipped without constructing any preference pairs.}

We run up to $T = 6$ iterations, producing models $M_1, M_2, \dots, M_6$. In each iteration, we train for one epoch on all the preference pairs constructed so far, with the number of pairs per iteration ranging from 10k to 30k, depending on the setting. \footnote{As model performance improves over iterations, fewer incorrect examples and more correct examples are generated, leading to varied number of preference pairs being constructed in each iteration.}

The loss coefficient $ \alpha $ is set to 0.5, and the DPO coefficient $ \beta $ is set to 0.4. Full-model fine-tuning is used, with a batch size of 8, gradient accumulation steps of 2, and a learning rate of $3 \times 10^{-8}$ using the AdamW optimizer with a constant learning rate schedule. Training is conducted on four A100 GPUs (80G memory) with a total batch size of 64.

\begin{figure*}[t]
\centering
\includegraphics[width=\textwidth]{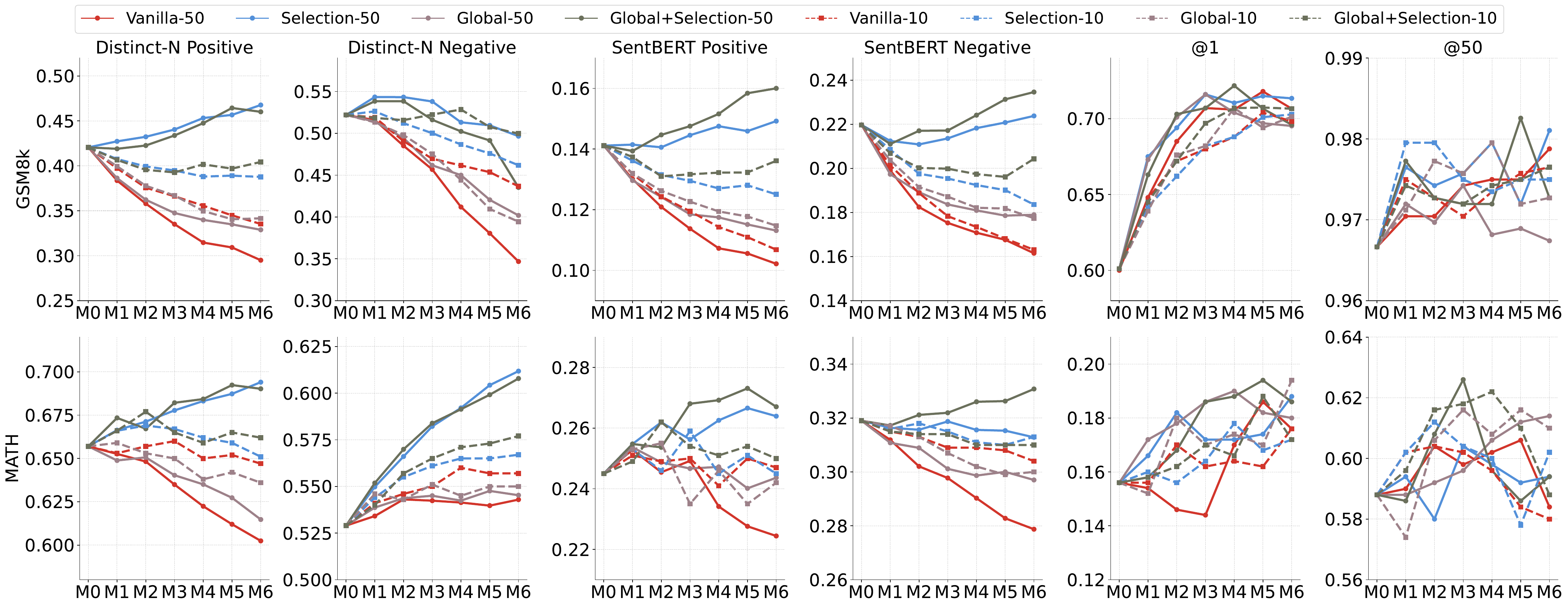}
\caption{Evolution of diversity metrics and model performance across iterations (M0-M6) for both GSM8k and MATH datasets. Each subplot shows different evaluation metrics: Distinct-N for positive and negative examples, SentBERT embeddings similarity, and accuracy measures. Solid and dashed lines with different colors represent different sampling settings and methods.} 
\label{fig: main_iterative_result}
\end{figure*}

\subsection{Experimental Results}
\label{sec:experimental-results}
To evaluate the effectiveness of our proposed methods, we conduct experiments with two sampling sizes (10 and 50) comparing four variants of ISI:

\begin{enumerate} 
\item Vanilla: The standard ISI method as our baseline
\item Global: Expanding sample pool with global data (Section \ref{sec:global-data})
\item Selection: Applying data selection for quality and diversity (Section \ref{sec:data-selection})
\item Global + Selection: Combining both global data expansion and data selection
\end{enumerate}

Tables \ref{tab: exp_result_gsm8k} and \ref{tab: exp_result_math} present the main results from the best-performing iteration (out of six) for each method. Our analysis reveals several key findings:

\textbf{Quality Preservation.} All three proposed methods (Global, Selection, and Global+Selection) maintain performance comparable to the baseline in terms of @1 and @50 accuracy on both GSM8K and Math datasets, demonstrating that our diversity-enhancing techniques do not compromise model quality.

\textbf{Impact of Sampling Pool Size.} With larger sampling size (50 vs 10), the vanilla method shows lower diversity, indicating that naive sampling expansion can actually harm diversity. Interestingly, the Global method alone does not consistently improve diversity over the vanilla baseline, suggesting that sample pool expansion without proper diversity management is insufficient.

\textbf{Effectiveness of Data Selection.} The data selection mechanism consistently enhances diversity across all settings (Sample 10/50, GSM8K/Math). This is evidenced by clear improvements from Vanilla to Selection and from Global to Global + Selection. Notably, the combination of large sampling (50) with Global + Selection achieves the highest diversity across most metrics.

\textbf{Iterative Analysis.} Figure \ref{fig: main_iterative_result} illustrates the dynamics across all six iterations:

1. \textbf{Diversity Evolution:} In vanilla ISI, diversity consistently declines across iterations, with larger sampling sizes (50) showing more severe reduction compared to smaller ones (10). Our Global + Selection method, in contrast, maintains and even improves diversity throughout iterations.

2. \textbf{Performance Trends:} All methods show accuracy improvements of 10-12 points on GSM8K and 2-4 points on Math, typically peaking between iterations 4-6 before saturation. The stable @50 accuracy across iterations suggests that self-improvement primarily acts as a re-ranking mechanism, consistent with observations in \citet{DBLP:journals/corr/abs-2407-05013}.

3. \textbf{Sample Size Effects:} Larger sampling (50) yields marginally better accuracy and significantly higher diversity compared to smaller sampling (10), indicating that increased sampling, when properly managed, benefits both quality and diversity.

\textbf{Ablation Analysis.} Our experiments serve as an ablation study to validate each component's contribution. For data selection, the consistent superiority of Selection over Vanilla in diversity metrics demonstrates its effectiveness. For sample pool expansion, the advantage of Global + Selection over Selection, larger sampling (50) over smaller sampling (10), confirms the benefit of incorporating global data. These results verify that both components are essential for maximizing diversity while maintaining performance.

\section{Analysis}
\begin{figure*}[t]
\centering
\includegraphics[width=\textwidth]{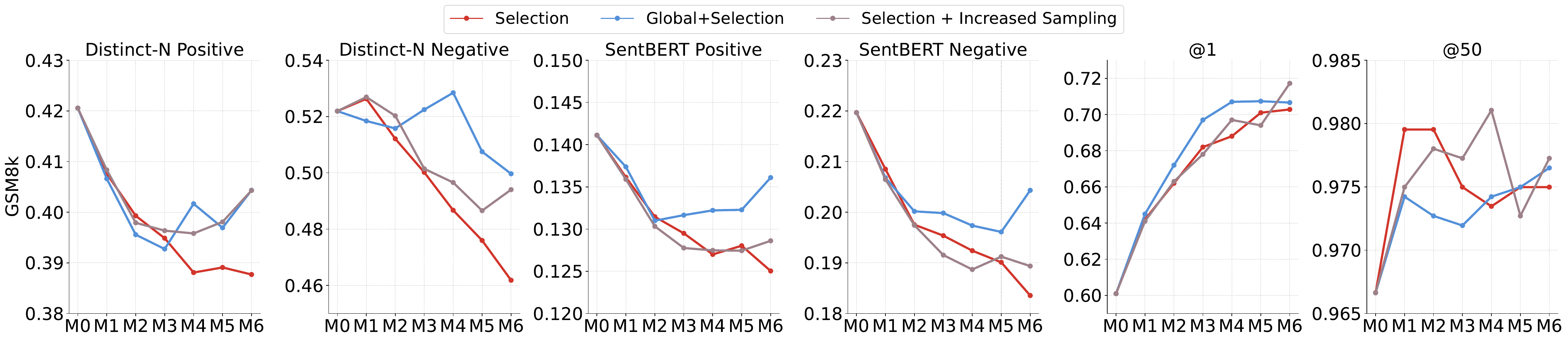}
\caption{Comparison of different sampling strategies for GSM8k dataset.} 
\label{fig: global_data_impact}
\end{figure*}

\begin{figure*}[t]
\centering
\includegraphics[width=\textwidth]{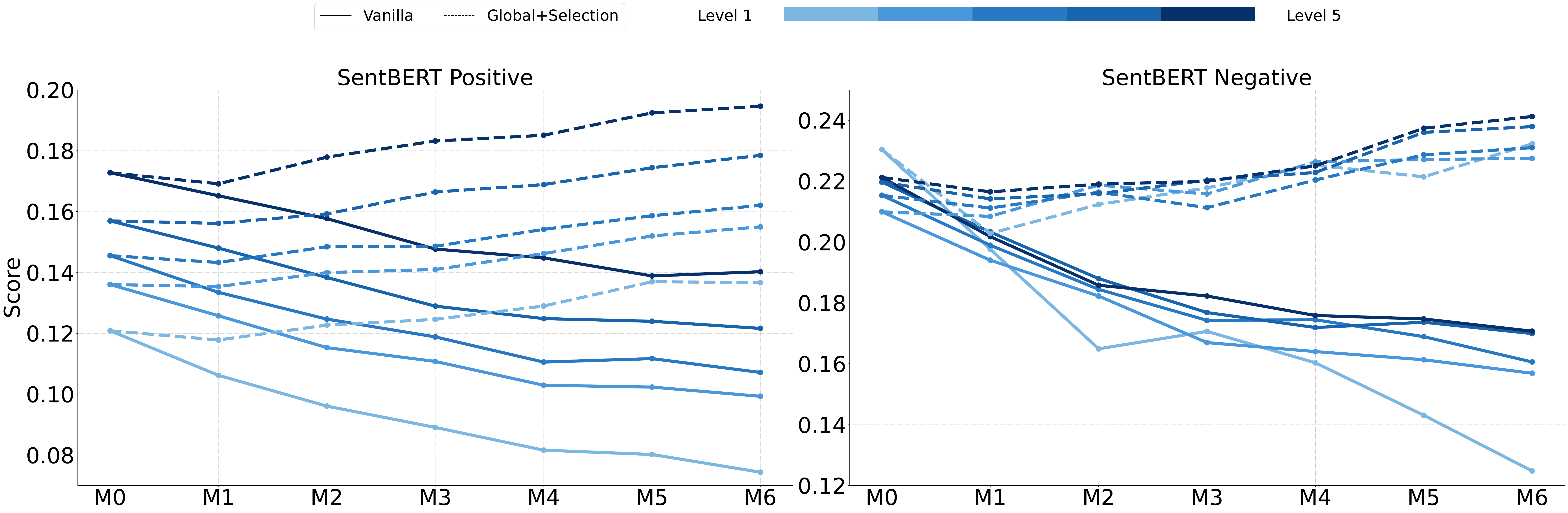}
\caption{Diversity trends across different difficulty levels (Level 1-5) for positive and negative examples. The plots demonstrate how question difficulty influences output diversity during the ISI process.} 
\label{fig: diff_level}
\end{figure*}

To gain deeper insights into diversity challenges in ISI and evaluate the effectiveness of \methodname, we investigate three key questions: Q1: Can increasing the number of samples per question alone adequately substitute for using a global data pool to expand the sample set? Q2: How does question difficulty affect diversity throughout the iterative process? Q3: How robust are our diversity improvements across different evaluation metrics?

\subsection{Impact of Global Data Usage (Q1)}
While both global data accumulation and increased per-question sampling can expand the sampling pool size, their effectiveness may differ. To investigate this, we compare three approaches across six iterations: 1.Selection: the sampling pool size remains constant at 10-10-10-10-10-10. 2.Global+Selection: the sampling pool size expands incrementally to 10-20-30-40-50-60 when global data is included, as each iteration incorporates all previous ones. 3.Selection+Increased Sampling: the sampling pool size is 10-20-30-40-50-60 via increased sampling count.

As shown in Figure \ref{fig: global_data_impact}, while Selection+Increased Sampling shows improved diversity in later iterations, Global+Selection consistently achieves higher diversity across all metrics for both positive and negative examples. This suggests that diversity lost in early iterations is difficult to recover through increased sampling alone, underscoring the importance of leveraging accumulated data. Moreover, Global+Selection achieves this with lower computational cost, requiring only 60 total samples per question compared to 210 for Selection+Increased Sampling, demonstrating both the effectiveness and efficiency of global data incorporation.

%\subsection{Diversity Across Difficulty Levels (Q2)}
%In Sec.\ref{fig: diff_level}, we conducted experiments on the GSM8K and Math datasets, observing that the Math dataset exhibits higher diversity than GSM8K, and the diversity loss during iterative self-improvement is less pronounced for Math. Since Math questions are more challenging than GSM8K, we hypothesize that question difficulty influences diversity in the iterative process.

%To test this, we followed \citep{} by sampling 50 examples per question and using the correct ratio as a proxy for difficulty level \footnote{Note correct ratio = R, Level 5: $0 \leq \text{R} < 0.2$; Level 4: $0.2 \leq \text{R} < 0.4$; Level 3: $0.4 \leq \text{R} < 0.6$; Level 2: $0.6 \leq \text{R} < 0.8$; Level 1: $0.8 \leq \text{R} < 1$}, allowing automatic difficulty classification without human annotation.

%Figure \ref{fig: diff_level} illustrates the diversity trends across difficulty levels. For positive examples, harder questions consistently show greater diversity, whereas this pattern is less pronounced for negative examples. During the iterative process, diversity decreases more significantly for easier questions (e.g., for negative examples, Level 1 drops by 53.4\%, Level 5 by 25.0\%; for positive examples, Level 1 drops by 43.8\%, Level 5 by 19.7\%). This aligns with the main experiment's finding that easier questions tend to have lower diversity and suffer greater diversity loss. Notably, DIVE effectively improves diversity across all difficulty levels without bias toward any specific level, confirming its robustness.

\subsection{Diversity Across Difficulty Levels (Q2)}
Our experiments on GSM8K and MATH datasets reveal an intriguing pattern: the more challenging MATH dataset maintains higher diversity and shows less pronounced diversity loss during ISI. This observation motivates us to investigate the relationship between question difficulty and diversity patterns.
To systematically analyze this relationship, we classify questions into five difficulty levels based on their correct ratio R (percentage of correct answers when sampling 50 examples)\footnote{Difficulty levels are defined as Level 5 (hardest): $0 \leq \text{R} < 0.2$; Level 4: $0.2 \leq \text{R} < 0.4$; Level 3: $0.4 \leq \text{R} < 0.6$; Level 2: $0.6 \leq \text{R} < 0.8$; Level 1 (easiest): $0.8 \leq \text{R} \leq 1$}. This automated approach enables objective difficulty assessment without manual annotation.

As shown in Figure \ref{fig: diff_level}, our analysis reveals several key findings: 1.Difficulty-Diversity Correlation: Harder questions consistently exhibit higher diversity in positive examples, though this correlation is less pronounced for negative examples. 2.Differential Diversity Loss: Easier questions suffer more severe diversity loss during iteration (e.g., Level 1 shows 53.4\% and 43.8\% drops for negative and positive examples respectively, compared to 25.0\% and 19.7\% for Level 5) 3.Method Robustness: DIVE demonstrates consistent diversity improvements across all difficulty levels, indicating its effectiveness is not biased toward any particular difficulty range.

\begin{figure*}[t]
\centering
\includegraphics[width=\linewidth]{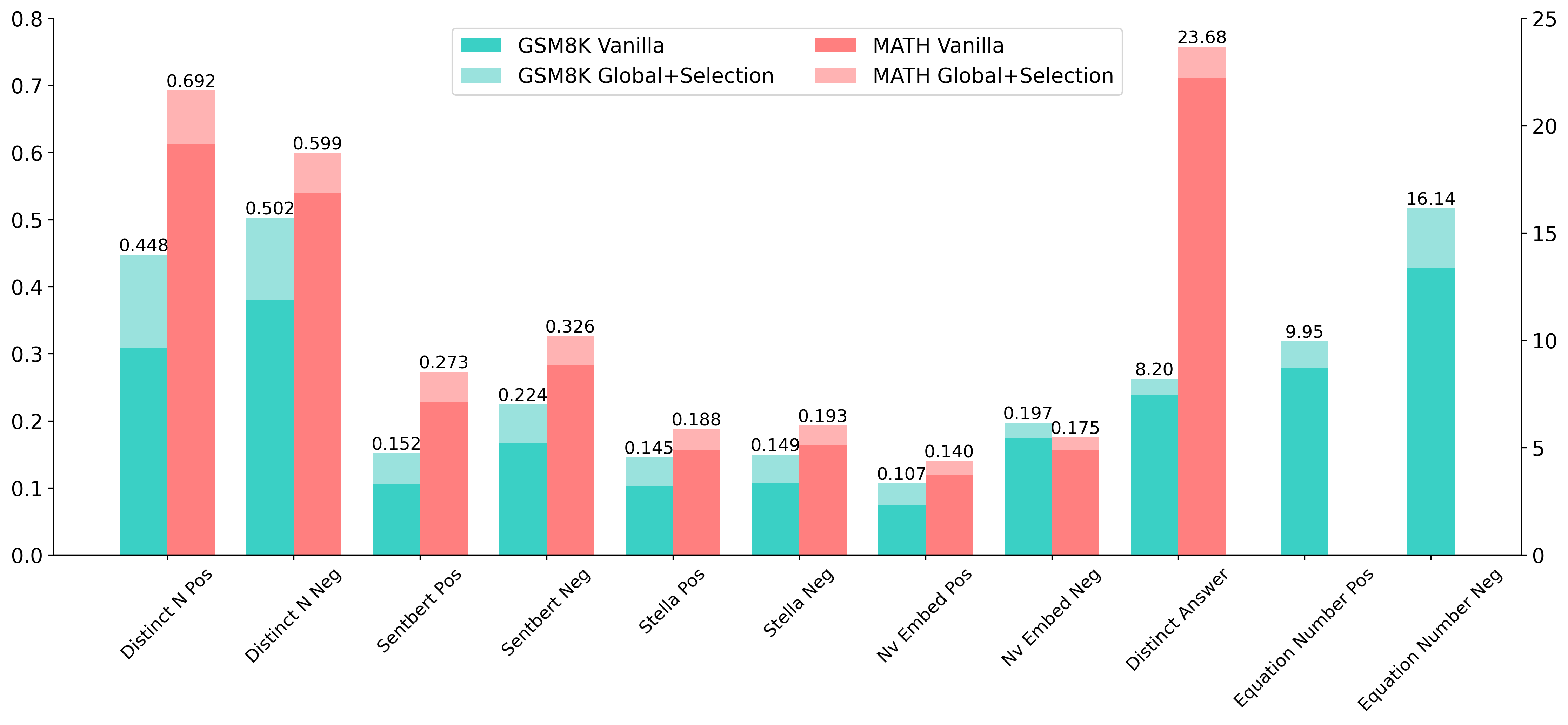}
\caption{Results of different diversity metrics for both the GSM8k and MATH datasets. Only the results from the iteration with the highest accuracy are shown, while the results for all iterations are provided in Appendix \ref{app: Iter_div}.} 
\label{fig: histogram_gsm8k_and_math}
\end{figure*}

\subsection{Alternative Metrics for Diversity (Q3)}\label{sec: analysis_Q3}
To validate the robustness of our diversity improvements, we extend our evaluation beyond the metrics in Section \ref{sec: evaluation metric}, incorporating both advanced embedding-based and task-specific metrics.

\paragraph{Advanced Embedding Metrics}
We employ two state-of-the-art embedding models for diversity assessment: \textbf{NV-Embed}~\citep{moreira2024nv}\footnote{Available at \url{https://huggingface.co/nvidia/NV-Embed-v2}}: A 7B parameter model currently leading the MTEB Leaderboard~\citep{muennighoff2022mteb}. \textbf{Stella}~\footnote{Available at \url{https://huggingface.co/dunzhang/stella_en_1.5B_v5}}: The top-performing 1.5B parameter model on MTEB.

\paragraph{Mathematical Reasoning Metrics}
We introduce two metrics specifically designed to capture diversity in mathematical reasoning: 
%\textbf{Equation Sequence Count}: Measures unique equation sequence patterns in solutions.\footnote{Applicable only to GSM8K due to its standardized equation notation using <<>>.} 
\textbf{Distinct Equation Chains}: This metric counts the number of distinct equation sequences in model-generated solutions, where each sequence represents a unique reasoning path.\footnote{This metric is only applicable to the GSM8K dataset due to its standardized equation notation using <<>>.}
\textbf{Distinct Answers}: Counts unique final answers per question, primarily reflecting diversity in incorrect solutions as correct answers are consistent.

As shown in Figure \ref{fig: histogram_gsm8k_and_math}, Global+Selection demonstrates consistent improvements across all eleven diversity metrics. Notably, while our method uses computationally efficient metrics (SentBERT and Distinct-N) during training, the improvements generalize to more sophisticated metrics, confirming the robustness of our approach.

\section{Related Work}
\paragraph{Diversity in Reasoning}
Research on diversity in language models has evolved from general text generation diversity~\citep{batra2012diverse,li2016simple,vijayakumar2018diverse} to the specific challenges of reasoning tasks, where the goal is to generate diverse yet valid solution paths. Recent work has explored various approaches: \citet{wang2022self} demonstrate that sampling multiple reasoning paths improves answer accuracy through aggregation, while \citet{xie2024self} combines beam search with temperature sampling to balance quality and diversity. Other approaches include varying prompts to enhance solution diversity~\citep{li2022making}, using model feedback to encourage multiple solving strategies~\citep{naik2023diversity}, and modeling reasoning as a Markovian flow for diverse path generation~\citep{yu2024flow}.

\paragraph{Iterative Self-Improvement}
Recent advances in ISI have shown promising results in enhancing model capabilities through self-play and iterative refinement, particularly in mathematical reasoning~\citep{DBLP:journals/corr/abs-2404-19733, mitra2024orca,DBLP:journals/corr/abs-2407-05013}. However, when models are trained on self-generated data, they may experience model collapse, where models progressively lose information about the underlying distribution~\citep{shumailov2024curserecursiontraininggenerated,dohmatob2024taletailsmodelcollapse,gerstgrasser2024modelcollapseinevitablebreaking}. This phenomenon has been observed in various settings including preference learning methods like DPO~\citep{rafailov2024direct} and RLHF~\citep{ouyang2022traininglanguagemodelsfollow}, where it manifests as reduced output diversity~\citep{kirk2023understanding, DBLP:journals/corr/abs-2407-05013}. While existing work suggests maintaining a balanced mix of human-authored and model-generated data to preserve model performance~\citep{shumailov2024curserecursiontraininggenerated, dohmatob2024taletailsmodelcollapse, gerstgrasser2024modelcollapseinevitablebreaking}, our work introduces a systematic approach to enhance diversity within the ISI framework itself.

\section{Conclusion}
We presented Diversified Iterative Self-Improvement (\methodname), a framework that addresses the challenge of diversity loss in ISI while maintaining model performance. Through systematic experiments on MATH and GSM8k datasets, we demonstrated that our two-component approach -- sample pool expansion and data selection -- effectively enhances output diversity across multiple evaluation metrics. Our experiments with different sampling sizes and detailed analysis across various difficulty levels demonstrated consistent improvements in diversity without compromising accuracy.

\section{Limitations}
While our work demonstrates the effectiveness of \methodname in mathematical reasoning tasks, several limitations should be noted:
\paragraph{Task Scope} Our study focuses exclusively on mathematical reasoning tasks (MATH and GSM8k). While we evaluate diversity using multiple metrics including equation patterns and embedding-based measures, the generalization of our approach to other domains remains to be explored.
\paragraph{Sampling Strategy} Although increasing the sampling size improves diversity, our current approach of fixed sampling per question may not be optimal. Questions of different difficulty levels might benefit from adaptive sampling strategies to better balance computational cost and diversity gains.
\paragraph{Computational Cost} Our experiments show that larger sample pools can enhance diversity, but the computational resources required increase significantly with sample size. While our global data usage method provides an efficient alternative to increased sampling, finding the optimal balance between pool size and computational cost remains a challenge.

\label{sec:appendix}
\begin{figure*}[t]
\centering
\includegraphics[width=\linewidth]{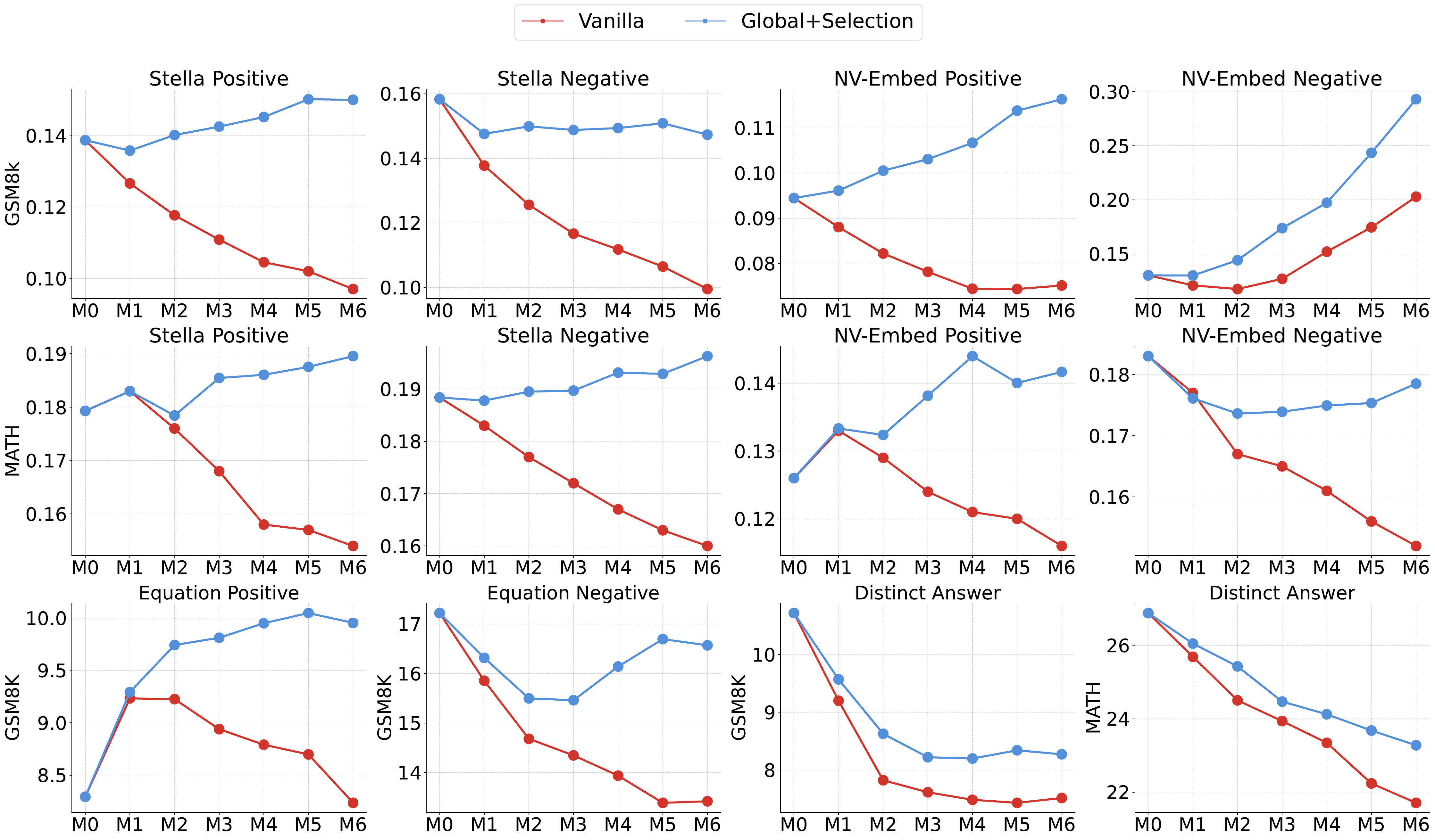}
\caption{Results of all iterations across different diversity metrics for both the GSM8k and MATH datasets.} 
\label{fig: appendix1}
\end{figure*}

% \section*{Acknowledgments}

\bibliography{main}

\appendix

\section{Appendix}

\subsection{Iterative Diversity Results by Alternative Diversity Metrics}\label{app: Iter_div}

Figure \ref{fig: appendix1} shows the full results of all iterations comparing the diversity of "Vanilla" and "Global+Selection" methods across six different diversity metrics, complementing the analysis in Section \ref{sec: analysis_Q3}. As seen, "Global+Selection" demonstrates higher diversity than "Vanilla" across all iterations and metrics. Moreover, the discrepancy increases with more iterations, highlighting the effectiveness of our method, particularly as the iterative process progresses.

\end{document}